\documentclass[lettersize,journal]{IEEEtran}

\usepackage[inline]{enumitem}

\usepackage[acronym]{glossaries} 
\usepackage{xspace} 

\usepackage{multirow}
\usepackage{tabulary}
\usepackage{booktabs} 

\usepackage{graphicx}
\usepackage[caption=false]{subfig}

\usepackage[many]{tcolorbox} 

\usepackage{color} 
\usepackage[ruled,vlined]{algorithm2e} 

\usepackage{bm} 
\usepackage[cal=boondoxo]{mathalfa} 
\usepackage{nicefrac} 

\usepackage{siunitx}

\usepackage{suffix} 
\usepackage{framed} 
\usepackage{xcolor} 
\usepackage[print-env=true]{scontents}
\newcommand{\seg}{$(ST)^2$\xspace}

\definecolor{main}{HTML}{5989cf}    
\definecolor{sub}{HTML}{cde4ff}     

\newtcolorbox{boxH}{
	width=0.9\linewidth,
	colback = sub, 
	colframe = main, 
	boxrule = 0pt, 
	leftrule = 4pt 
}

\newcommand{\quoteUser}[3]{%
	\begin{flushright}%
	\begin{boxH}
			\emph{``#3''}
			\begin{flushright}
				User \##1 regarding the #2 method.
			\end{flushright}%
	\end{boxH}%
	\end{flushright}%
}

\newcommand{\gpFont}[1]{\mathcal{#1}} 

\DeclareMathOperator*{\argmax}{arg\,max}

\newcommand{\dataset}{\gpFont{D}}
\newcommand{\policy}{\pi}
\newcommand{\policyPos}{p}
\newcommand{\policyTime}{t}
\newcommand{\numDemos}{n}
\newcommand{\demoNumSamples}{M}
\newcommand{\demo}{\tau}
\newcommand{\numSegments}{s}
\newcommand{\segmentNumSamples}{r}

\newcommand{\covTrainTrain}{\gpFont{K}}
\newcommand{\covPointPoint}{\gpFont{k}}
\newcommand{\covPointTrain}{\covPointPoint_\star}

\newcommand{\gpBaseKernel}{\tilde{\covPointPoint}}
\newcommand{\gpState}{\gpFont{x}}
\newcommand{\gpStates}{\gpFont{X}}
\newcommand{\gpLabel}{\gpFont{y}}
\newcommand{\gpLabels}{\gpFont{Y}}
\newcommand{\gpMean}{\mu}
\newcommand{\gpVar}{\sigma}

\newcommand{\stiffness}{K}
\newcommand{\stiffnessSaturated}{\tilde{K}}
\newcommand{\varThreshold}{\gpVar_{th}}
\newcommand{\damping}{D}

\newcommand{\jacobian}{\nabla J}
\newcommand{\torqueCommands}{\rho}

\newcommand{\smdpNumTasks}{m}
\newcommand{\smdpFulltask}{\gpFont{T}}
\newcommand{\smdpSubtask}{T}
\newcommand{\smdpAction}{o}
\newcommand{\smdpSubgoal}{g}
\newcommand{\smdpFinalgoal}{\smdpSubgoal_\smdpNumTasks}
\newcommand{\smdpTerminationFn}{\beta}

\newcommand{\sttCommand}{c}
\newcommand{\sttCommandAuto}{\sttCommand_{\rm{auto}}}
\newcommand{\sttCommandNext}{\sttCommand_{\rm{next}}}
\newcommand{\sttCommandDemo}{\sttCommand_{\rm{demo}}}
\newcommand{\sttCommandInsertKP}{\sttCommand_{\rm{insertKP}}}
\newcommand{\sttCommandReset}{\sttCommand_{\rm{reset}}}
\newcommand{\sttPolicies}{\Pi}
\newcommand{\sttMode}{\rm{mode}}

\renewenvironment{quote}{\begin{fquote}\itshape\advance\leftmargini -2.4em\begin{oldquote}}{\end{oldquote}\end{fquote}}

\newenvironment{fquote}
{%
	\MakeFramed {\advance\hsize-3\width \FrameRestore}
}
{\endMakeFramed}

\long\def\RC#1\par{\setlength{\parskip}{0.6\baselineskip}\setlength{\parindent}{0pt}\makebox[0pt][r]{\bf RC:\hspace{4mm}}\textbf{{#1}}\par} 
\WithSuffix\long\def\RC*#1\par{\textbf{{#1}}\par} 

\long\def\AR#1\par{\setlength{\parskip}{0.6\baselineskip}\setlength{\parindent}{0pt}\makebox[0pt][r]{AR:\hspace{10pt}}{#1}\par} 
\WithSuffix\long\def\AR*#1\par{{#1}\par} 
\newacronym{bparhmm}{BP-AR-HMM}{Beta-Process AutoRegressive Hidden Markov Model}

\newacronym{cnn}{CNN}{Convolutional Neural Networks}
\newacronym{cic}{CIC}{Cartesian Impedance Controller}

\newacronym[longplural={Dynamical Systems}]{ds}{DS}{Dynamical System}
\newacronym{dmp}{DMP}{Dynamic Movement Primitive}
\newacronym[longplural={Degrees of Freedom}]{dof}{DoF}{Degree of Freedom}
\newacronym{dl}{DL}{Deep Learning}

\newacronym{fmea}{FMEA}{Failure Mode and Effects Analysis}

\newacronym[longplural={Gausian Processes}]{gp}{GP}{Gaussian Process}
\newacronym[longplural={Graph Gausian Processes}]{ggp}{GGP}{Graph Gaussian Process}
\newacronym{gmm}{GMM}{Gaussian Mixture Model}
\newacronym{gpr}{GPR}{Gaussian Process Regression}

\newacronym{iil}{IIL}{Interactive Imitation Learning}

\newacronym{hmm}{HMM}{Hidden Markov Model}
\newacronym{hsmm}{HSMM}{Hidden Semi-Markov Model}
\newacronym{hrc}{HRC}{Human-Robot Collaboration}
\newacronym{hrec}{HREC}{Human Research Ethics Committee}

\newacronym{kt}{KT}{kinesthetic teaching}
\newacronym{kd}{KD}{kinesthetic demonstration}

\newacronym{lfd}{LfD}{Learning from Demonstrations}
\newacronym{llm}{LLM}{Large Language Model}
\newacronym{lstm}{LSTM}{Long Short-Term Memory}

\newacronym{pca}{PCA}{Principal Component Analysis}

\newacronym{mp}{MP}{Movement Primitive}
\newacronym{mdp}{MDP}{Markovian Decision Process}

\newacronym{nasa-tlx}{NASA-TLX}{NASA Task Load Index}
\newacronym{nn}{NN}{Neural Network}
\newacronym{nlp}{NLP}{Natural Language Processing}

\newacronym{rl}{RL}{Reinforcement Learning}
\newacronym{rbf}{RBF}{Radial Basis Function}

\newacronym{simple}{SIMPLe}{Safe Interactive Movement Primitive Learning}
\newacronym{smdp}{SMDP}{Semi-Markov Decision Process}

\newacronym{tphsmm}{TPHSMM}{Task Parameterized Hidden Semi-Markov Model}
\newacronym{tpgmm}{TPGMM}{Task Parameterized Gaussian Mixture Model}
\newacronym{tamp}{TAMP}{Task and Motion Planning}

\begin{document}


\title{Sequentially Teaching Sequential Tasks \seg:\\ 
Teaching Robots Long-horizon Manipulation Skills}


\author{
    Zlatan Ajanović\IEEEauthorrefmark{1}, Ravi Prakash\IEEEauthorrefmark{1}, Leandro de Souza Rosa\IEEEauthorrefmark{1}, Jens Kober
    \thanks{Z. Ajanovic is with the Computer Science Department, RWTH Aachen University, 52062 Aachen, Germany (e-mail: zlatan.ajanovic@ml.rwth-aachen.de).}
    \thanks{R. Prakash is with Cyber Physical Systems, Indian Institute of Science Bengaluru, India (e-mail: ravipr@iisc.ac.in).}
    \thanks{L. de Souza Rosa is with Alma Mater Studiorum Università di Bologna, Italy (e-mail: leandro.desouzarosa@unibo.it).}
    \thanks{J. Kober is with the Cognitive Robotics Department, Delft University of Technology, 2628 CD Delft, The Netherlands (e-mail: J.Kober@tudelft.nl).}
    \thanks{\IEEEauthorrefmark{1} Equal contribution. Authors were affiliated with TU Delft at the time of the study.}
}



\maketitle

\begin{abstract}
Learning from demonstration has proved itself useful for teaching robots complex skills with high sample efficiency.
However, teaching long-horizon tasks with multiple skills is challenging as deviations tend to accumulate, the distributional shift becomes more evident, and human teachers become fatigued over time, thereby increasing the likelihood of failure.
%
%
To address these challenges, we introduce \seg, a sequential method for learning long-horizon manipulation tasks that allows users to control the teaching flow by specifying key points, enabling structured and incremental demonstrations. Using this framework, we study how users respond to two teaching paradigms: (i) a traditional monolithic approach, in which users demonstrate the entire task trajectory at once, and (ii) a sequential approach, in which the task is segmented and demonstrated step by step.
We conducted an extensive user study on the restocking task with $16$ participants in a realistic retail store environment, evaluating the user preferences and effectiveness of the methods.
User-level analysis showed superior performance for the sequential approach in most cases (10 users), compared with the monolithic approach (5 users), with one tie. Our subjective results indicate that some teachers prefer sequential teaching---as it allows them to teach complicated tasks iteratively---or others prefer teaching in one go due to its simplicity.
\end{abstract}


\section{Introduction}\label{sec: introduction}

\gls{lfd} has emerged as an effective paradigm for teaching robots complex manipulation skills with minimal data and high sample efficiency.
By leveraging expert demonstrations, robots can bypass low-level motion planning and focus on replicating high-level behaviors.
However, when scaling to long-horizon tasks composed of multiple sequential or interdependent sub-tasks, \gls{lfd} methods face significant challenges.
Small deviations from the demonstration can accumulate over time, leading to cascading errors and distributional shifts that degrade policy performance.

\begin{figure}[t]
	\centering
	\includegraphics[width=1\linewidth, trim=0 80 150 80, clip]{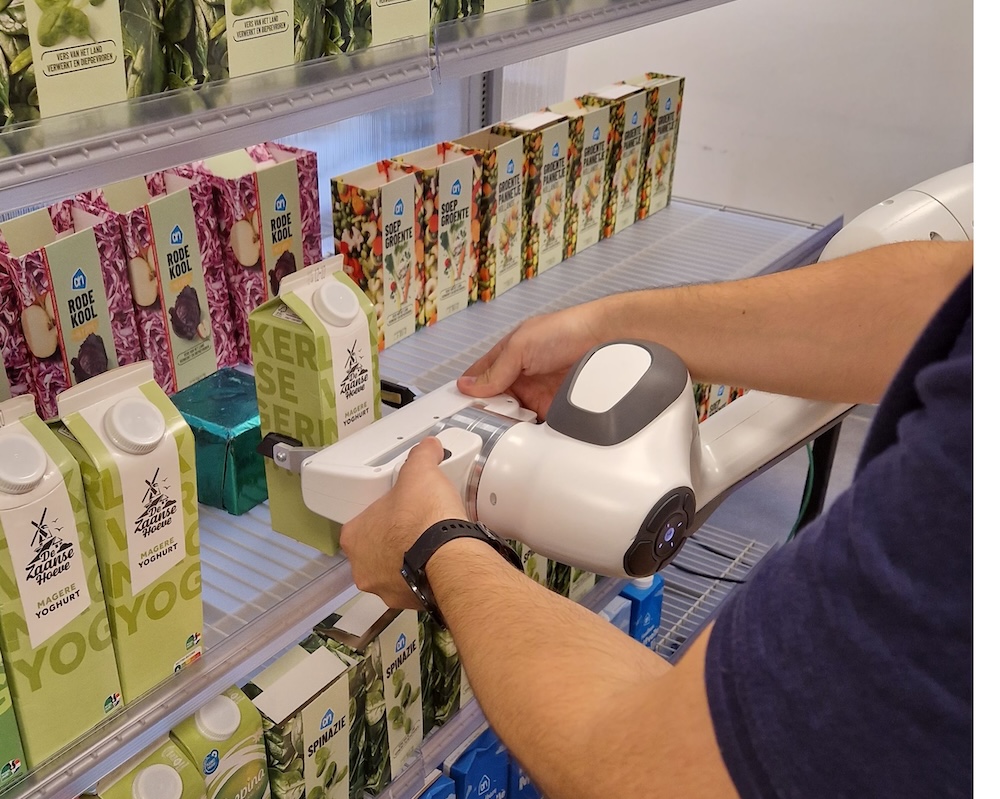}
	\caption{
		Teaching a supermarket restocking task.
	}
	\label{fig:eyecatcher}
\end{figure}

Moreover, \gls{kt}, the most common demonstration process due to its efficiency and performance \cite{jiang2024comprehensive}, becomes increasingly demanding for the teacher as it is physically tiring for human instructors, particularly for long-horizon tasks, e.g., the supermarket restocking task illustrated in Figure \ref{fig:eyecatcher}.
In this setup, the quality and consistency of demonstrations often decline as fatigue and the mental demand of thinking about several steps while providing demonstrations accumulate, deprecating the data quality.
These limitations highlight the need for approaches that can structure, segment, and compose demonstrations in ways that support robust generalization over long task horizons.

Despite the progress on \gls{lfd} and \gls{iil} methods, the targeted tasks are usually simple and performed in a single motion \cite{celemin2022interactive}, and extensions to long-horizon tasks have been tackled by several methods in the literature by ordering sub-tasks, or skills, to perform a complex task.
As such, these methods assume a partition, or segmentation, of demonstrations into smaller tasks.

Towards assembling a long-horizon task from segmented demonstrations, early approaches \cite{perez2017c} focus on sequencing sub-tasks using key-frames, which are important states shared among executions of the same task.
Naturally, ordering non-sequential segments has also been explored, e.g., in \cite{mandlekar2020learning}, a low and a high-level policy are used to learn sub-tasks and their coordination, respectively.
Recently, approaches aim at improving generalization, e.g., \cite{patton2024programming} uses demonstrations partitioned into an ``alphabet'' of sub-tasks, from which regular expressions are inferred to learn a program encompassing all demonstrations, making it possible to combine skills as a form of generalization to new tasks.


Alternatively, long-horizon tasks can be approached using \gls{tamp} methods in synergy with \gls{lfd}.
As an example, \cite{welschehold2019combined} builds a likelihood model encoding the goal and teacher's intention, allowing to learn a task-tree used in planning new tasks by optimizing the co-occurrence of objects and the teacher's hands.


Regarding the data acquisition process, \cite{mandlekar2023human} frames long-horizon task learning with the \gls{tamp} framework, enabling the composition of complex tasks from simpler ones; it also allows for switching control to the human teacher when needed, and input information to be used for multiple sub-tasks, reducing the human-time requirements.


From the literature, it becomes clear that segmentation methods play a fundamental role in learning long-horizon tasks, since sub-tasks encoding the segments allow learning higher-level models easily by offloading the low-level training.
This idea has been explored in \cite{li_hierarchical_2021}, where pre-defined segments are identified on the demonstrations and used for learning sequential and hierarchical policies with reduced complexity.
%

To create segments from full demonstrations provided by the user, a common approach is to use \glspl{hmm}-based methods, an idea first introduced in \cite{fox_bayesian_2009}, where a non-parametric model handles multi-scale segmentation due to its hierarchical auto-regressive nature. 
Bayesian inference has been used for segmentation and estimating their number automatically based on velocity profiles \cite{gutzeit2022unsupervised}.
Furthermore, \cite{figueroa2018physically} considers the physical constraints of robotic manipulator trajectories to create an asymptotically stable dynamical system encoding the desired robot behavior.
Nevertheless, these methods require the user to provide full demonstrations beforehand, without addressing the challenges that they impose on the human teachers.

When learning from a human teacher, there are two important aspects to consider.
The first, and the focus of the literature mentioned above, is the learning efficiency due to the quality and scarcity of demonstrations that human teachers can produce and how they affect the policy learning \cite{mandlekar2021matters}.
The second are human aspects impacting the teaching, e.g., comfort, preferences, and safety, during the training sessions, which require user studies to evaluate.

For \gls{hrc} tasks, \cite{tausch2022best} evaluates the effects of task allocation, showing that operators reported greater satisfaction and production results improved whenever users could design their task sequence.
Furthermore, \cite{pantano2023effects} compares expert and novice teachers regarding their preferences for teamwork and effort, showing that expert users prioritize efficiency, while novice ones value comfort.
Therefore, evaluating the user's preferences regarding the training framework is an important aspect in \gls{lfd} and \gls{iil}.

In this paper, we focus on evaluating the human teacher's preferences when teaching long-horizon tasks in two frameworks: \textbf{i)} the traditional full task demonstration, hereafter named ``monolithic''; \textbf{ii)} a sequential teaching method where teachers define segments according to their intuition. 

Our key contributions are:
\begin{enumerate*}
	\item \seg, a novel method for teaching long-horizon sequential manipulation tasks that allows the teacher to control teaching flow and teaching sub-tasks incrementally (sequentially).
	\item A benchmark task with an analysis compatible with \gls{fmea} \cite{fmea} (from the production engineering field), which allows for comparing both teaching frameworks directly.
	\item An extensive user-study and analysis with $16$ participants, yielding insights on human preferences in teaching and resulting policy quality.
\end{enumerate*}
\section{Background}\label{sec: background}

In this section, we provide the necessary background for our study, including the imitation learning approach adopted in this work, the kinesthetic teaching interface, and a problem formulation for long-horizon manipulation tasks.

\subsection{\acrfull{simple}}\label{subsec: simple}

The backbone method for our study is the \gls{simple} framework \cite{franzese2023interactive}, which encodes full demonstrations using a time-dependent \glspl{gp}-based policy, providing a prediction of epistemic uncertainty that can be used for disturbance rejection or stiffness regulation \cite{franzese2021ilosa}.

In \gls{simple}, $\numDemos$ different demonstrations $\demo$ are recorded forming a dataset $\dataset = \{\demo_i\}_{i=1}^\numDemos$, each demonstration being composed of a sequence of $\demoNumSamples_i$ states $\demo_i = \{\gpState_i^j\}_{j=1}^{\demoNumSamples_i}$, which in turn represent the robot's state $\policyPos$ (end-effector's Cartesian pose and grip state), and the sample's timestamp $\policyTime$, i.e., $\gpState_i^j = (\policyPos_i^j, \policyTime_i^j)$.
The labels for each sample are taken as the next point\footnote{The last point's label is itself.} in the same demonstration sequence, $\gpLabel_i^j = \gpState_i^{j+1}$.

The states and labels are concatenated into the $\gpStates$ and $\gpLabels$ sets, respectively, and a policy $\policy$ is learned using a \gls{gp}, providing goal predictions for the robot given its current $\gpState$.
Particularly, the goal is given by the \gls{gp}'s prediction mean and the variance provides an epistemic uncertainty estimation at the evaluation point, both computed as described in Equation~\eqref{eq: GP}, where the $\covPointPoint=\covPointPoint(\gpState,\gpState)$ is the current state's variance $\gpState\notin\gpStates$, $\covPointTrain=\covPointPoint(\gpStates,\gpState)$ is the $\gpState$ and training inputs $\gpStates$ variance, and $\covTrainTrain=\covTrainTrain(\gpStates,\gpStates)$ is the covariance matrix of the $\gpStates$ \cite{williams2006gaussian}.  

\begin{equation}\label{eq: GP}
	\begin{array}{rll}		
		\gpMean(\gpState) & = &  \covPointTrain^\top\covTrainTrain^{-1} \gpLabels \\
		\gpVar(\gpState) & = & \covPointPoint -\covPointTrain^\top\covTrainTrain^{-1}\covPointTrain
	\end{array} 
\end{equation}

A key aspect of \gls{simple} is its time encoding in the \gls{gp}'s state, which evolves the policy as it progresses during task execution to provide superior encoding capabilities for complex trajectories, e.g., when similar states are visited multiple times, creating ambiguous estimations otherwise.
To cope with the dynamic policies, \gls{simple} employs the pseudo-\gls{gp} kernel which correlates states only with their closest neighbors, as described in Equation \eqref{eq: GP kernel}, where  $\gpBaseKernel=e^{-|\policyPos_i-\policyPos_j|\lambda^{-1}}e^{-|\policyTime_i-\policyTime_j|\lambda^{-1}}$ is a composed exponential kernel for the robot's pose $\policyPos$ and time $\policyTime$.

\begin{equation}\label{eq: GP kernel}
	\covPointPoint(\gpState_i,\gpState_j)= 
	\begin{cases}
		1 ,& \text{if } \gpState_j = \displaystyle\argmax_{\gpState_k\in\gpStates}\gpBaseKernel(\gpState_i, \gpState_k)\\
		0,              & \text{otherwise}
	\end{cases}
\end{equation}

The correlation approximation in Equation \eqref{eq: GP kernel} turns $\covTrainTrain$ into an identity matrix, avoiding the computationally expensive $\covTrainTrain^{-1}$ in Equation \eqref{eq: GP}, which usually leads to prohibitively low control frequencies.
Furthermore, when using $\covPointPoint$, the \gls{gp} returns the closest point label as prediction, which can be interpreted as a graph, leading to the proposed name \gls{ggp}.

\gls{simple} employs two mechanisms to ensure safe and smooth operation, which are not guaranteed given the \gls{gp} approximation.
The first being an adequate sampling rate ($>$\qty{10}{\hertz}) to guarantee state coverage and the \gls{ggp}'s approximation stability.
Second, it employs a \gls{cic} with dynamically regulated stiffness ($\stiffness$) and damping ($\damping$) matrices as described in Equation \eqref{eq: cic}, where $\torqueCommands$ is the actuation values of the robot's controller, and $\jacobian$ is the robot's Jacobian.
$\stiffness$, $\damping$ are dynamically computed using the current state's uncertainty $\gpVar(\gpState)$ as $\stiffness=\stiffnessSaturated\nicefrac{(1-\gpVar(\gpState))}{(1-\varThreshold)}$, with $\varThreshold$ being a variance threshold, $\stiffnessSaturated$ a diagonal matrix with values limited to respect user-given maximum velocity and force thresholds, and $\damping=2\stiffness^{\nicefrac{1}{2}}$.
As such, when uncertainty increases towards $\varThreshold$, the $\stiffness$ automatically drops, stalling the robot's automatic operation and passing control back to the human teacher, who will provide feedback in the form of corrections used to refine the policy $\policy$.

\begin{equation}\label{eq: cic}
	\torqueCommands = \jacobian\left[\stiffness\left(\gpMean(\gpState)-\gpState\right)-\damping\left(\dot{\policyPos}_r\right)\right]
\end{equation}

\subsection{Teaching Interface and Corrections}

The primary interface for providing demonstrations in this study is the \glsreset{kt}\gls{kt}, enabled by making the robot physically compliant, with its joints relaxed and easy to move, and in gravity compensation mode.
From a control perspective, this is achieved by setting the robot's \glsreset{cic}\gls{cic} to have very low $\stiffness$ and $\damping$ values, allowing the human teacher to grasp the robot's arm and guide its end-effector smoothly through the desired path.
As such, \gls{kt} allows non-expert users to easily teach robots new movements by guiding them without needing any programming knowledge.

As the user moves the robot, \gls{simple} continuously records the robot's state $\gpState$, including its spatial orientation and position ($\policyPos$) and the timestamp ($\policyTime$) at which that pose was reached, forming the demonstration trajectory $\demo$ which composes the states and labels ($\gpStates$ and $\gpLabels$) used to train the policy $\policy$.

Note that in our \gls{iil} setup, users provide $\numDemos=1$ demonstrations to start up the policy, which can be iteratively refined through corrections.
To do so, the robot starts performing the task controlled by the initial policy.
If the robot deviates from the expected trajectory or stalls in regions of high uncertainty, the teacher can grasp the robot and move it to the desired pose.
Such situations can be easily identified by the presence of external forces from the robot's joint-torque sensors, triggering \gls{simple} to start recording the newly visited states $\gpState$, which are appended to $\gpStates$ and $\gpLabels$, allowing for incorporating new information for the following run by simply retraining $\policy$. 
Furthermore, note that corrections are local, meaning that slightly displacing the robot to a well-defined region in the policy is sufficient to correct its policy, and the robot will automatically restart performing the task.

Besides \gls{kt}, we allow users to provide high-level spoken language commands that steer the learning process and arbitrate autonomy between robot and human.

\subsection{A Problem of Teaching Long-Horizon Manipulation Tasks}\label{subsec: smdp}

We consider the problem of teaching long-horizon manipulation tasks 
$$
\smdpFulltask = \{\smdpSubtask_1, \smdpSubtask_2, \dots, \smdpSubtask_\smdpNumTasks\},
$$ 
where each sub-task $\smdpSubtask_i$ corresponds to achieving a sub-goal $\smdpSubgoal_i$ 
(e.g., top grasping, collision-free reaching, placing, gripper release), and the overall execution of $\smdpFulltask$ achieves the final goal $\smdpFinalgoal$.
We assume that demonstrations are provided via \gls{kt}, yet monolithic trajectories are prone to error accumulation and distributional shift, while also causing user fatigue.
To mitigate this, the teacher should be able to dynamically control the teaching flow, switching between demonstration, autonomous execution, and correction, thereby enabling localized error recovery at the sub-goal level.
Formally, this can be viewed as a simplified \gls{smdp} \cite{sutton1999mdps} with \emph{options} representing sub-tasks $\smdpSubtask_i \in \smdpFulltask$, where the challenge lies in learning consistent policies $\smdpAction_i = (\policy_i, \smdpTerminationFn_i)$ for achieving sub-goals \(\{\smdpSubgoal_i\}\), while ensuring smooth transitions and maintaining usability for the human teacher.

\section{Methodology}\label{sec: methodology}

Our goal is to compare two \gls{iil} frameworks for teaching robotic manipulation tasks from the teacher's perspective.
The first framework is the standard \gls{iil} one, hereinafter named ``monolithic teaching'', in which the user initially provides a demonstration of the entire task.
In the second framework, users are allowed to demonstrate the task as a series of segments encoding ``actions'' by setting key-points on-the-fly and to control their creation and order with high-level user commands, thus enabling \emph{progressive skill acquisition}, and \emph{adaptive interaction} between human teachers and robots.
In both cases, the user can provide real-time corrections during task execution, allowing for policy refinement.

To test the latter, we propose a novel method called \seg, which implements a \gls{simple} policy for each user-defined segment.
The intuition behind \seg's user-centered segmentation is to balance instructional complexity with the teacher's capacity, mitigating cognitive overload while avoiding potential misalignment and ensuring the systematic buildup of skills.
	
\subsection{Sequentially Teaching Sequential Tasks \seg}\label{subsec: st2}

Our teaching framework introduces a structured approach to robot learning through demonstration, featuring explicit control over when demonstrations are provided, policies are executed, and key-points are inserted (skills are segmented). This framework enables incremental task learning and supports fine-grained user intervention during the teaching process.

Since we consider the demonstration to be divided into $\numSegments$ segments, the dataset becomes the set of demonstrations for each segment $\dataset = \{\demo_i\}_{i=1}^{\numSegments}$, with $\demo=\{\gpState_i^j\}_{j=1}^{\segmentNumSamples_i}$, where $\segmentNumSamples_i$ is the number of samples in the segment $i$.
The segments are used to learn a \gls{simple} policy for each segment in $\dataset$, forming the set $\sttPolicies=\{\policy_i\}_{i=1}^{\numSegments}$.

Regarding the teaching flow, the teacher can perform five high-level commands to control the teaching process, summarized in Equation \eqref{eq: commands}, giving the user control over task segmentation by allowing the insertion of key-points and switching between demonstration and autonomous execution intuitively.

\begin{equation}\label{eq: commands}
	\sttCommand \in 
	\begin{cases} 
	    \sttCommandAuto: & \text{executes full task autonomously } \\
	    \sttCommandDemo: & \text{allows user to provide demonstration} \\
	    \sttCommandInsertKP: & \text{inserts key-point and starts new segment} \\
	    \sttCommandNext: & \text{executes action until the next key-point} \\
	    \sttCommandReset: & \text{resets to the episode's start }
	\end{cases}
\end{equation}

Algorithm \ref{alg: st2} outlines the sequential teaching framework designed for interactive and incremental policy learning.
The process begins by initializing an empty dataset $\dataset$ and operating in demonstration gathering mode.
At each time step, the system either records the robot’s current pose (gathering data during demonstration) or autonomously executes the next action according to the current policy $\policy_i$ (during autonomous execution).
The teacher can dynamically influence the learning process through the high-level $\sttCommand$ commands: 
The command $\sttCommandDemo$ sets the system to gather a segment demonstration, and $\sttCommandInsertKP$ inserts a new key-point that ends the current segment demonstration and initializes a new segment demonstration.
The command $\sttCommandReset$ resets the environment and robot state to their initial conditions, and $\sttCommandNext$ runs the current policy until the subsequent key-point, and new points are added to the trajectory in case the user decides to correct the policy.
After each iteration, policies are retrained on the updated dataset $\mathcal{D}$, ensuring continuous refinement as new data becomes available.
For testing, $\sttCommandAuto$ sets the system to autonomous mode, in which the complete task is performed by sequentially executing the policies for each segment.

We highlight that the teacher is capable of overwriting segments by autonomously executing the previous actions and providing a new demonstration for the target segment and the subsequent ones, hence editing the segmentation's granularity.
To do so, the user can autonomously execute the autonomous policy of segments that perform their respective sub-tasks successfully, and then take over the control to provide new demonstrations and key-points. Note that handling the transition to the newly presented demonstrations is unnecessary, since they start from the previous segment’s ending point.
To highlight, this feature differs from \gls{simple}'s local corrections, which can be used to fine-tune existing policies with kinesthetic feedback while executing autonomously.

\begin{algorithm}
    \caption{Sequential Teaching Framework \seg}
    \label{alg: st2}
    \SetKwInOut{Parameters}{Parameters}
    \KwOut{Updated dataset $\dataset$, set of learned policies $\sttPolicies = \{\policy_i\}$}
    
    $\dataset \gets \emptyset$ \tcp{Initialize dataset}
    $i \gets 1$ \tcp{Initialize key-point index}
    $\demo_i \gets \emptyset$ \tcp{Initialize segment data}
    $\policyTime \gets 1$ \tcp{Initialize time}
    $\sttMode \gets \sttCommandDemo$ \tcp{Start in demo mode}
    
    \While{\texttt{True}}{
        $\gpState \gets [\texttt{get\_pose}(), \policyTime]$\\
        \uIf{$\sttMode == \mathtt{demo}$}{
            $\demo_i \gets \demo_i \cup \gpState$ \tcp{Append data}
            $\policyTime \gets \policyTime + 1$\\
            \If{$\sttCommandInsertKP$}{
                $\dataset \gets \dataset \cup \demo_i$ \tcp{Add segment to dataset}
                $i \gets i + 1$ \tcp{Insert new keypoint}
            	$\policyTime \gets 1$\\
            	$\demo_i \gets \emptyset$ \tcp{Start new segment}
            }
        }
        \ElseIf{$\sttMode == \mathtt{auto}$}{
            \If{$\policyPos \approx \policyPos_i^{\segmentNumSamples}$}{
                $\sttMode \gets \mathtt{pause}$ \tcp{Reached termination condition}
            }
            \Else{
            	\texttt{execute($\policy_i(\gpState)$)}\\
            	$\policyTime \gets \policyTime + 1$\\
            	\If{user\_input()}{
            		$\demo_i \gets \demo_i \cup \gpState$
            	}
            }
        }
        
        \uIf{$\sttCommandDemo$}{
            $\sttMode \gets \texttt{demo}$ \tcp{Switch to demonstration}
        }
        \ElseIf{$\sttCommandNext$}{
            $\sttMode \gets \texttt{auto}$ \tcp{Mode is autonomous}
            $i \gets i + 1$ \tcp{Move to next segment}
            $\policyTime \gets 1$
        }
        \ElseIf{$\sttCommandReset$}{
            $\mathtt{goto}(\policyPos_1^1$) \tcp{Reset to initial state}
            $i \gets 1$ \\
            $\policyTime \gets 1$
        }
        $\sttPolicies \gets \texttt{train}(\dataset)$ \tcp{Retrain policies}
    }
\end{algorithm}

\begin{scontents}[store-env=smooth transitions]
	As a final remark, note that \gls{simple}'s time encoding and graph-like correlation ensure that corrections will be close to the current evaluation point, and its safety mechanisms ensure smooth transitions between different segments.
	Therefore, no additional mechanisms are required to ensure smooth and safe trajectories.
\end{scontents}
\section{Experiment}\label{sec: experiment}

This section details the experimental setup designed to evaluate two imitation learning strategies (monolithic and sequential teaching) through a structured robotic manipulation task involving human participants.

\subsection{Supermarket restocking task}\label{subsec: restocking task}

For our user study, we have chosen the real-world supermarket restocking problem, in which a robot transfers a milk/yogurt carton from a storage box to a shelf, reflecting a practical scenario where robots assist in logistics and retail operations.
This task presents several challenges that necessitate a multi-step approach due to the spatial constraints imposed by the shelf design, specifically the inability to place the carton in an upper shelf while holding it from the top.
As the robot can grasp the carton from its box only from its top (due to the box sides), the robot must pick up the carton, place it on a table, and re-grasp it from the side before moving it to the shelf.
One potential segmentation of this task into $10$ sub-tasks is shown in Table \ref{table:tasks}.

\begin{figure*}
	\centering
	\includegraphics[width=1\linewidth]{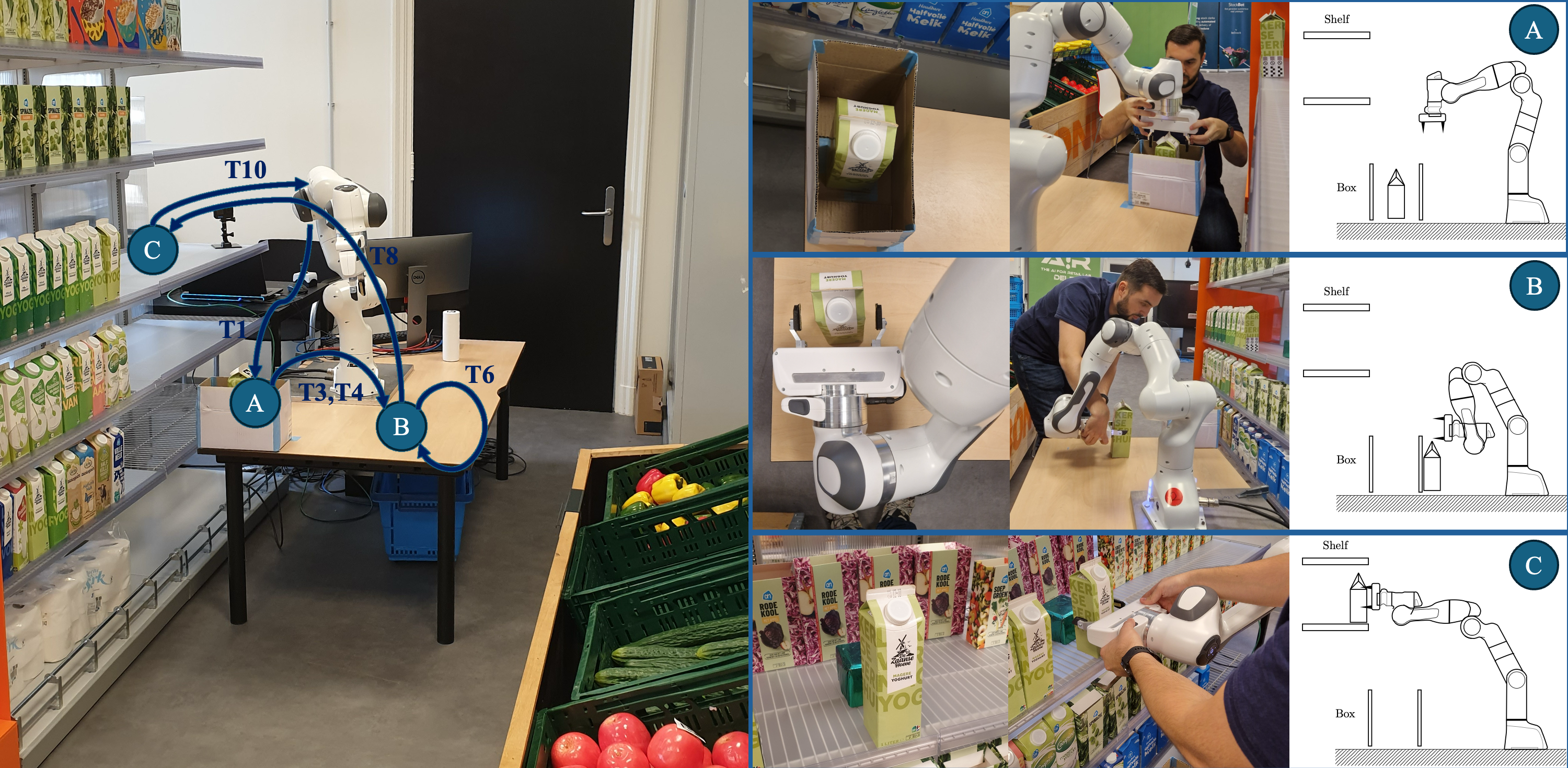}
	\caption{
		Restocking task. A) Initial state, carton in the box, B) Intermediate placement for re-grasping, C) Final state, carton on the shelf.
	}
	\label{fig:task_main}
\end{figure*}

The supermarket restocking task was chosen for the user study as it embodies key characteristics essential for evaluating robotic manipulation and human teaching strategies.
The task's moderate complexity ensures that it is neither trivial nor overwhelmingly difficult, requiring the robot to perform sequential, long-horizon actions such as re-grasping due to spatial constraints.
This structure allows for clear observability and measurability, as success can be quantified through task completion time, grasp efficiency, and placement accuracy.
Additionally, the task ensures safety and reproducibility, as it involves non-hazardous objects and can be consistently replicated across trials.
Importantly, it offers high training value, providing a structured example of learning multi-step manipulation skills while maintaining a manageable cognitive load for human demonstrators.

\begin{table}[t]
	\centering
	\caption{Task Breakdown for Robotic Manipulation}
	\label{table:tasks}
	\begin{tabulary}{\linewidth}{C|L|C}
		\toprule
		\textbf{Task ID} & \textbf{Description} & \textbf{Component} \\ \midrule
		T1 & Move to reach carton & arm \\ \hline
		T2 & Top grasp carton & gripper \\ \hline
		T3 & Move out of the box & arm \\ \hline
		T4 & Move above the location on the table & arm \\ \hline
		T5 & Release the grasp & gripper \\ \hline
		T6 & Reposition end-effector for side grasp & arm \\ \hline
		T7 & Side grasp carton & gripper \\ \hline
		T8 & Move to the shelf & arm \\ \hline
		T9 & Release the object & gripper \\ \hline
		T10 & Move out of the shelf & arm \\ \bottomrule
	\end{tabulary}
\end{table}

Since we are interested in studying the differences in teaching a monolithic and segmented policy from the human teacher's perspective, the supermarket restocking task is adequate as it provides the necessary complexity to make these differences apparent.
Nevertheless, other tasks can be considered for extensions of this study, reinforcing our findings or providing new insights.

\subsection{Problem Complexity and Task Analysis}\label{subsec: problem complexity}

The supermarket restocking task has been selected through \gls{fmea} principles \cite{fmea}, a structured methodology for evaluating task complexity, identifying potential failure points, and assessing overall system performance, crucial aspects to \gls{iil}, where ensuring reliability is critical.

Regarding the task's complexity, we first decompose it to its minimum granularity, identifying individual steps involved in its execution, as summarized in Table \ref{table:tasks}.
For each step, Table \ref{table:fmea} presents a summary of the identified potential failures, such as inaccuracies in motion replication, incorrect mental models, or hardware limitations, and their potential effects on performance, safety, and efficiency, enabling us to pinpoint critical vulnerabilities in both the learning and execution phases.
Although the restocking problem might seem simplistic, it is clear that it is challenging, and many errors can occur in practice.

\begin{table*}[ht]
	\centering
	\caption{Errors and Potential Effects in Robotic Manipulation for Supermarket Restocking}
	\label{table:fmea}
	\begin{tabulary}{\textwidth}{C|C|L|L}
		\toprule
		\textbf{Category} & \textbf{Code} & \textbf{Error Name} & \textbf{Unwanted Effect} \\ \midrule
		\multirow{9}{*}{Teaching} 
		 & ET01 & Object moved by hand or falls & Missing object location at execution \\ \cline{2-4} 
		 & ET02 & Object moved within the gripper & Missing object exact pose  \\ \cline{2-4} 
		 & ET03 & Moving close to other objects & Execution not robust, touching other objects \\ \cline{2-4} 
		 & ET04 & Touching other objects while moving & Execution not robust, pushing the other objects \\ \cline{2-4} 
		 & ET05 & Moving the arm while grasping or releasing & Grasping execution not robust \\ \cline{2-4} 
		 & ET06 & Moving back and forth while demonstrating & Robot behavior not smooth \\ \cline{2-4} 
		 & ET07 & Reaching joint limits while demonstrating & Robot behavior not smooth  \\ \cline{2-4} 
		 & ET08 & Non-perpendicular object placing & Placing execution not robust \\ \cline{2-4} 
		 & ET09 & Non-perpendicular object placing grasping (e.g., pushing it while grasping) & Picking execution not robust \\ \hline
		\multirow{2}{*}{Task Planning} 
		 & EP01 & Grasping carton from the top and trying to place it directly (skipped steps) & Teaching failure \\ \cline{2-4} 
		 & EP02 & Placing object at reachability limit (e.g., far placement, joint limits) & Impossible to continue the task \\ \hline
		\multirow{2}{*}{Segmenting} 
		 & ES01 & Not enough segments (far less than tasks) & $(ST)^2$ method not useful for corrections \\ \cline{2-4} 
		 & ES02 & Meaningless key-frames & $(ST)^2$ method not useful for corrections \\ \hline
		\multirow{2}{*}{Correcting} 
		 & EC01 & Missing the desired time to start correction & Missed opportunity to correct \\ \cline{2-4} 
		 & EC02 & Forgetting to use segmentation in followup demonstrations when correcting & $(ST)^2$ method not useful for correction\\\hline
		 Setup & EG00 & Different causes for gripper malfunctioning & Gripper does not perform taught action \\
		 \bottomrule
	\end{tabulary}
\end{table*}

Regarding user performance, Section \ref{sec: results} presents results that evaluate task execution using objective and subjective metrics, user feedback, and teaching preferences.

As a final remark, perception is not considered in this study since the underlying learning method would require several demonstrations for learning generalized policies (e.g., \cite{franzese2025generalizable}), greatly increasing the mental demand on the human teachers and the number of possible failure scenarios, hindering the user-preference analysis.

\subsubsection{Monolithic Policy Teaching}\label{subsubsec: complexity monolithic}

In the monolithic policy teaching approach, the entire complex task is demonstrated to the robot as a single, uninterrupted movement.
Demonstrating a task that involves multiple steps from start to finish in one go implies that the natural task breakdown into multiple steps must be executed fluently, which we extrapolate to be more difficult for inexperienced human teachers.

On the learning side, movements with small pose variations, e.g., during re-grasping, yield ambiguous points, ultimately leading to the learned police to either interpolate the movements to their average or falling into a deadlock.
Even though this problem is mitigated by \gls{simple}'s time encoding, we argue that segmenting the task helps reduce potential ambiguous states created in these situations.

\subsubsection{Sequential Policy Teaching}\label{subsubsec: complexity seg}

Besides reducing the policy learning complexity, the proposed segmentation has the advantage of keeping the semantic meaning intended by the users, which we extrapolate to be helpful to them, improving their capabilities of providing useful corrections and hence benefiting the learning process.
	
Nevertheless, a generic learning framework that considers segments would also require learning a full \gls{smdp}, a problem we avoid with the simplifications made in Section \ref{subsec: smdp}, given that our goal is to evaluate the differences from the human teacher's perspective.

\subsection{User Study}\label{subsec: user study}

The user study\footnote{Approved by the TU Delft \gls{hrec}.} starts with participants reading and signing the provided information and informed consent sheets.
The study is divided into three main sessions: Familiarization, Experimental, and Questionnaire.

During the \emph{Familiarization Session} (\qty{15}{\minute}), participants receive instructions and feedback to familiarize themselves with the robot and \gls{kt} setup, which is done by moving the robot in compliant mode to explore different joints, understanding the robot's joints limits, avoiding collisions, practicing opening and closing the gripper, and reorienting the gripper to pick objects from various orientations and positions.
Additionally, participants practice teaching a simple pick-and-place sequence using both the monolithic and sequential methods.

In the \emph{Experimental Session} (\qty{30}{\minute}), participants evaluate the two demonstration methods by manipulating the robot arm to perform a restocking task described in Section \ref{subsec: restocking task} using the monolithic and \seg frameworks, in random order to avoid learning bias.
Note that the task partition presented in Table \ref{table:tasks} is not shown to the participants, who have total autonomy to segment the task according to their own understanding and preferences.

Due to \gls{simple}'s encoding capabilities, a single initial demonstration and subsequent corrections are sufficient for learning a successful policy in most cases.
All \gls{simple} policies were trained in a few seconds on a standard laptop computer (with an \texttt{Intel Core 7} processor, without the use of a GPU) due to the \gls{gp}'s kernel approximation, rendering policy training time negligible in this study.
Users can repeat the complete learning process from scratch up to $3$ times for each method, capped by the elapsed time (\qty{1}{\hour}) and the user's level of fatigue.
    
During the \emph{Questionnaire Session} (\qty{20}{\minute}), participants provide their opinions and preferences on each demonstration method by completing the \gls{nasa-tlx} questionnaire, a widely used subjective assessment tool to rate human perceived workload and performance for a given task.
Participants also answer additional open-ended questions regarding their experiences and preferences for the demonstration methods.

\subsubsection*{Research questions}

This study explores the following key questions:

\begin{itemize} 
	\item \textbf{RQ1:} Is the \seg (sequential)  method more user-friendly or effective than the monolithic method? 
	\item \textbf{RQ2:} Under what conditions does one teaching method become preferable over the other?
\end{itemize}

We hypothesize that users who successfully demonstrate the entire task in a single attempt may prefer the monolithic method due to its simplicity and speed.
However, when the robot fails to reproduce the demonstrated behavior reliably, users may shift their preference toward the sequential approach, which offers greater control and opportunities for correction. 
\section{Results and Discussion}\label{sec: results}

Our user study counts $16$ participants, $3$ female and $13$ male, ranging from $22$ to $40$ years old.
Among them, $14$ had no experience with \gls{lfd}, and $9$ had never interacted with a robot before the study.
A total of $6$ participants were familiar with robotic manipulators.
As such, the study is performed with users with diverse expertise levels, which is important since in \gls{lfd} setups the learning performance is greatly impacted by the teacher's experience \cite{mandlekar2021matters}.

\begin{figure}[h]
	\centering
	\includegraphics[width=\linewidth]{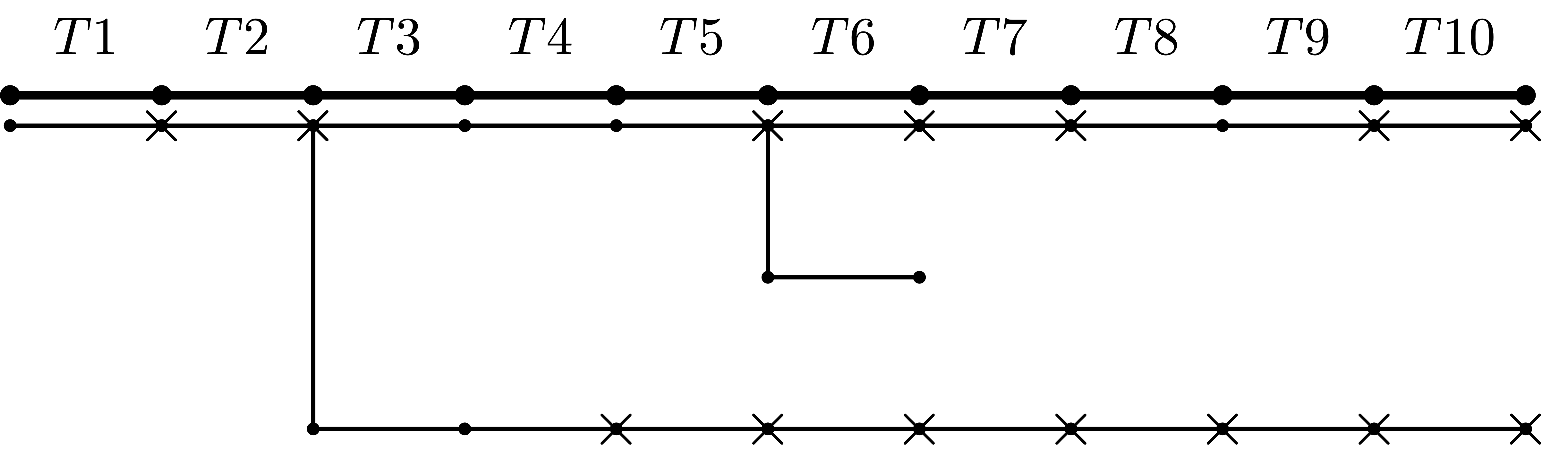}
	\caption{User $\#03$ \seg teaching flow.}
	\label{fig: teaching flow}
\end{figure}

%
To illustrate the teaching flow and how autonomy switches between robot and user, while the latter provides demonstrations and segmentation key-points, we utilize the rooted branching tree representation exemplified in Figure \ref{fig: teaching flow}.
The top line (main) represents the complete task $\smdpFulltask$, with all sub-tasks, dots represent the corresponding keypoints at the end of the sub-tasks presented in Table \ref{table:tasks}, and user-provided key-points are marked with a cross.
Below the main line, the rooted branching tree represents all demonstrations in a single teaching trial; branches are created when the user requests autonomous execution until a certain point, and provides new demonstrations from that point onward.
In this example, the user segments the demonstration with $7$ keypoints (skipping $\smdpSubtask_3$, $\smdpSubtask_4$, and $\smdpSubtask_8$).
The user wanted to improve and requested autonomous execution until $\smdpSubtask_5$, but gave up soon because keypoints were lacking, and the robot went too far from the expected trajectory, making the user request a reset.
In the second correction run, the user executed autonomously until $\smdpSubtask_2$ and continued providing demonstrations and key-points until the end.
In some examples, users had more than $4$ correction rounds, progressively advancing the success of sub-task execution.

\subsection{User Preferences}\label{subsec: user preferences}

\begin{figure}
	\centering
 	\includegraphics[width=0.8\linewidth]{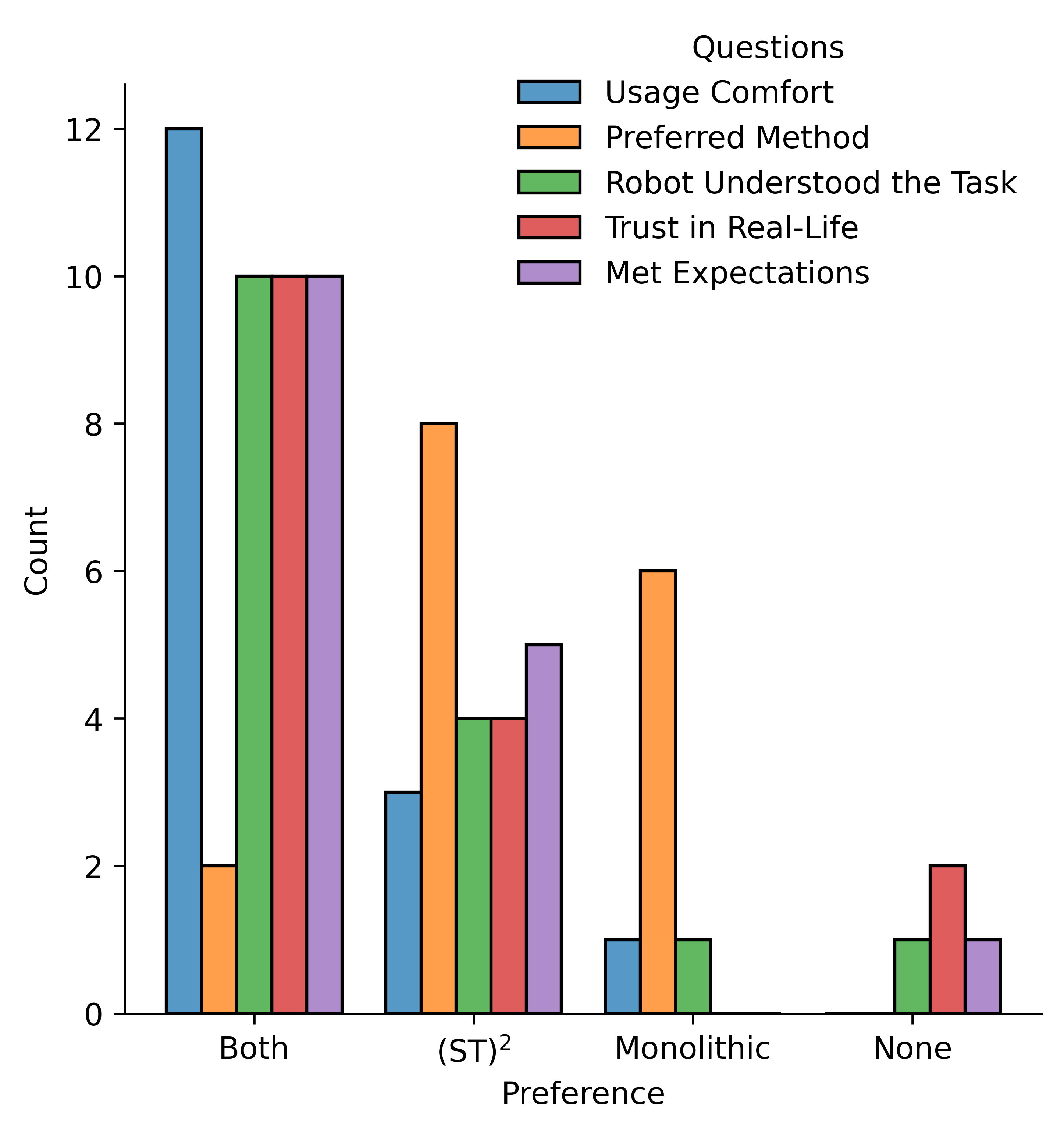}
 	\caption{Distribution of users' preferences and subjective evaluation regarding the learning methods.}
 	\label{fig: quantitative}
\end{figure}

Users were asked about their preference between teaching with the monolithic method or with \seg regarding their comfort, trust, expectations, and their perception of the robot's understanding of the task.
Figure \ref{fig: quantitative} summarizes the users' answers. While there is a clear distinction in their preference (i.e., they either prefer the monolithic or sequential method), most users declared being comfortable and trusting both methods, indicating that there is no downfall in teaching the robot in either way.
 
Furthermore, a significant portion of users favor \seg regarding trust, expectation, and their perception of the robot's understanding.

\subsection{Subjective User Evaluation}\label{subsec: subjective user evaluation}

\begin{figure*}
 	\centering
 	\subfloat[Users who preferred monolithic teaching.]{
		\includegraphics[width=1.0\linewidth]{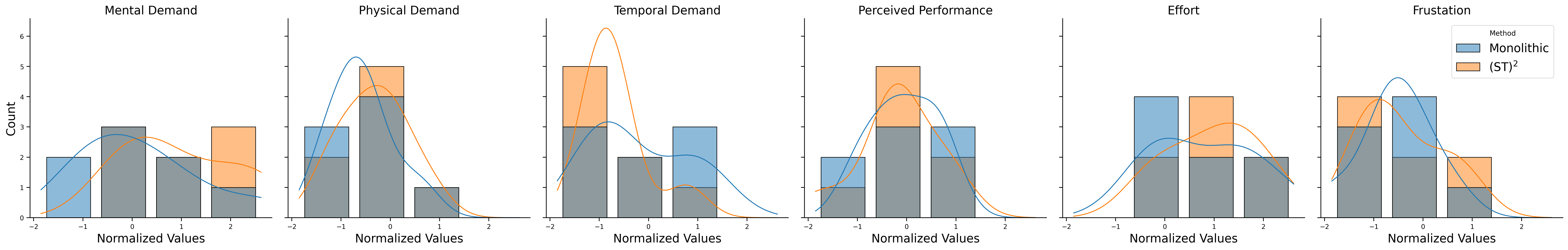}
 	}%
 	\\
 		\subfloat[Users who preferred \seg.]{
 		\includegraphics[width=1.0\linewidth]{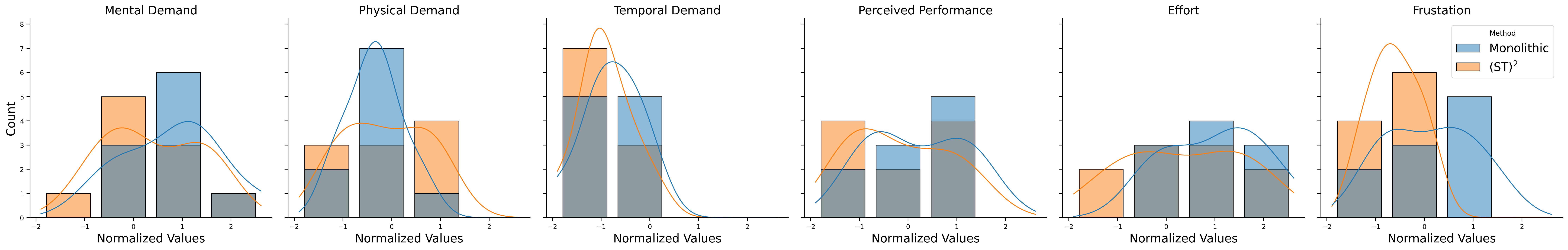}
 	}%
 	\caption{
 		Quantitative \gls{nasa-tlx} workloads normalized (Z-score) per user. Smaller is better. Answers are divided according to the user's preferred method to highlight how the workloads affect their preferences. Users who preferred both methods are counted in both figures.
 		}
 	\label{fig:workloads}
\end{figure*}

Figure \ref{fig:workloads} presents the \gls{nasa-tlx} answers highlighting the users' subjective assessment of the tasks divided by the users' preferred method; $6$, $8$, and $2$ participants preferred the monolithic, \seg, and both methods, respectively.
Results show a smaller physical demand using the monolithic method (overall $\rm{p-value}=0.01$), which we attribute to its ``in one-go'' demonstration being less physically demanding, as pointed out by users.
\quoteUser{03}{monolithic}{
	It felt more satisfying getting the robot to perform the complete “choreography” correctly in a single movement.
}

Interestingly, \seg is perceived as the less temporally demanding method (overall $\rm{p-value}=0.03$), which is caused by users investing a significant amount of time correcting the monolithic policy.
Despite \gls{simple} allowing for local corrections, finding the correct starting and stopping times for the corrective feedback proved challenging for non-expert users, who would continue to provide corrections until the task's end, as explicitly pointed out by five users in their comments.
\quoteUser{16}{monolithic}{
	Potentially, I would have to re-demonstrate everything to the robot if it fails.
}

An interesting point is that one could expect a higher mental demand when using \seg, since users have to think about key-points.
In fact, $3$ users reported difficulties in deciding when to create a key-point.
However, results show no significant difference in the mental demand (overall $\rm{p-value}=0.42$).
Upon further inspection, since \seg follows a natural way of teaching tasks, the expected mental demand overhead was commonly overcome, as pointed out by $5$ users.

\quoteUser{02}{\seg}{
	This task consists of sub-tasks. So I think I am more comfortable as a demonstrator to achieve the final goal if I can teach tasks separately. Also, I do not need to consider the optimization of the whole task, which reduces the need to think.
}

\quoteUser{07}{\seg}{
	Yes, it allowed me to ``explain'' the task in segments, which is also how I prefer to do demonstrations to humans. Also, a small mistake felt less ``stressful''.
}

Finally, we observe no statistical differences in the performance perceived by the users, nor in their effort and frustration self-evaluations.

\subsection{Task Performance}\label{subsec: task performance}

\begin{figure}
	\centering
	\includegraphics[width=0.9\linewidth]{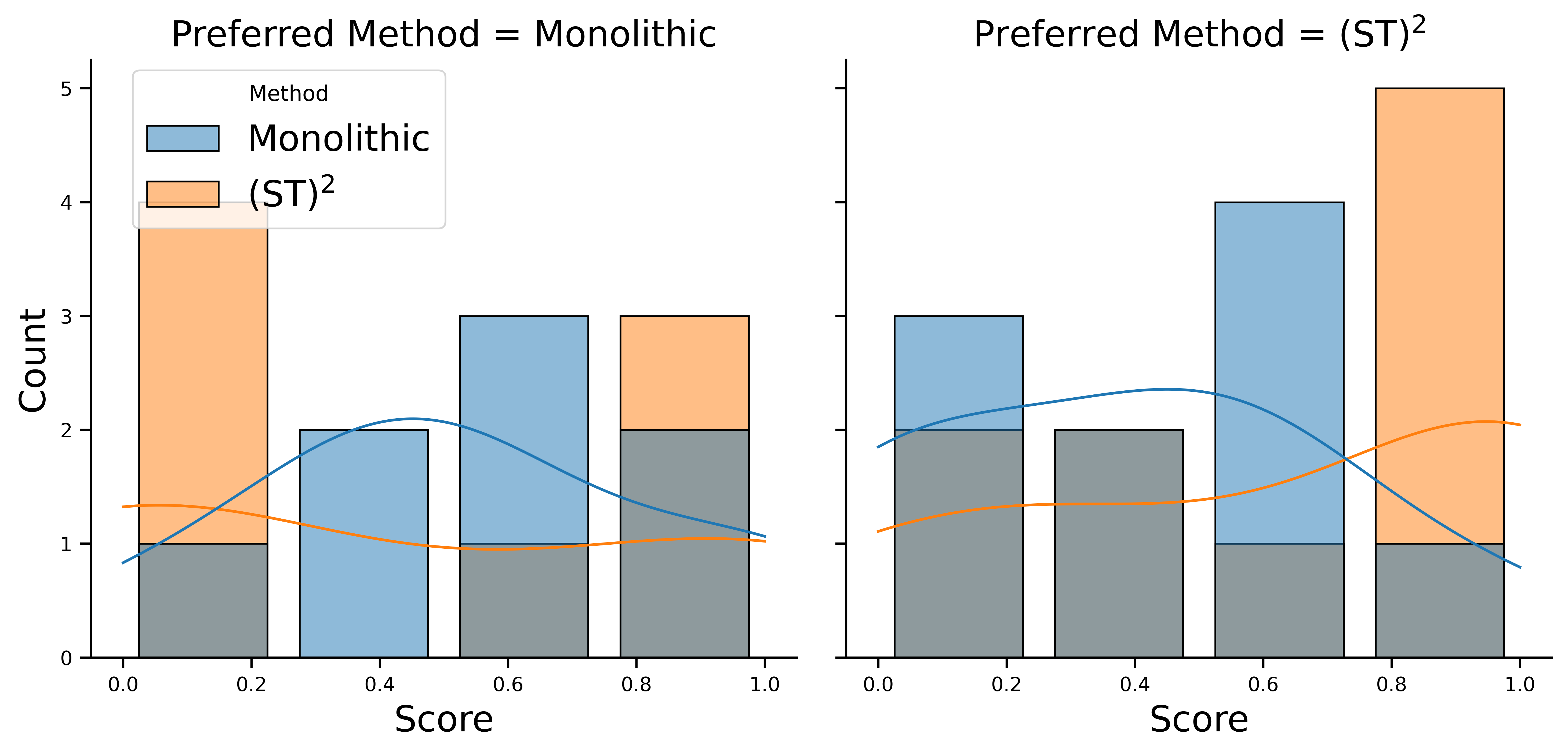}
	\caption{Task performance score distribution measured $score=\nicefrac{1}{\#t}$, where $\#t$ indicates number of necessary trials for a successful execution. Answers are divided according to the user's preferred method, and users who preferred both methods are counted in both figures. Higher is better.}
	\label{fig: scores}
\end{figure}

Figure \ref{fig: scores} presents the overall task scores for training the robot using both approaches and split by user preference.
Despite a subtle alignment between users preferring \seg and their low score using the monolithic method, when evaluating all users altogether, we do not observe a significant score difference between methods ($\rm{p-value} = 0.98$), with success rates for the monolithic and \seg being $51.1$ and $65.5$, respectively. User-level analysis revealed that users improved performance with the sequential approach in most cases (10 users), compared with the monolithic approach (5 users), with one tie.
	
We speculate these results are related to the teaching flow difference between methods.
Since the monolithic method requires users to time their corrections when the robot diverges from the expected path, users see it as a full demonstration, distracting them from the possibility of correcting locally.
On the other hand, \seg allows for focusing corrections only on problematic segments, reinforcing the feeling of providing local corrections, as pointed out by $3$ users.
\quoteUser{05}{\seg}{
	I prefer the \seg method because I was able to correct just after a certain step. If everything was good up to that point, I could ``save my work''.
}
As a consequence, the focus on specific segments when using \seg allowed users to provide meaningful corrections only when needed, resulting in a successful task with fewer runs w.r.t. the monolithic method.

\subsection{Quality Metrics}\label{subsec: quality metrics}

Table \ref{tab: quality metrics} provides a quality comparison between the trajectories generated by policies learned with both methods.
The trajectory and torque jerk results show that \seg generates smoother trajectories, while also successfully performing the task with similar speed but a higher success rate.
Such results provide indicative evidence that our assumptions about the user being able to provide better demonstrations and corrections when he/she has control over the segmentation culminate in smoother and high-performing policies.

\begin{table}[h]
    \centering
    \caption{Comparison of Frobenius Norms of Jerk Product Matrices (Mean $\pm$ Std. Dev.) and average policy time to complete the task successfully.}
    \label{tab: quality metrics}
    \begin{tabulary}{\linewidth}{c|CCC}
        \toprule
         & \textbf{Trajectory Jerk ($10^{4}$)} & \textbf{Torque Jerk ($10^{10}$)} & \textbf{Execution Time} (s) \\
        \midrule
        \textbf{Monolithic} & $10.5  \pm 9.99 $ & $21.1  \pm 32.5 $ & $46.1 \pm 11.5$  \\
        \textbf{\seg} & $8.91  \pm 3.05 $ & $7.54  \pm 4.11 $ & $46.3 \pm 11.8$ \\
        \bottomrule
    \end{tabulary}
\end{table}

\subsection{Learned Behavior}

\begin{figure}%
	\centering
	\subfloat{\includegraphics[width=0.4\textwidth]{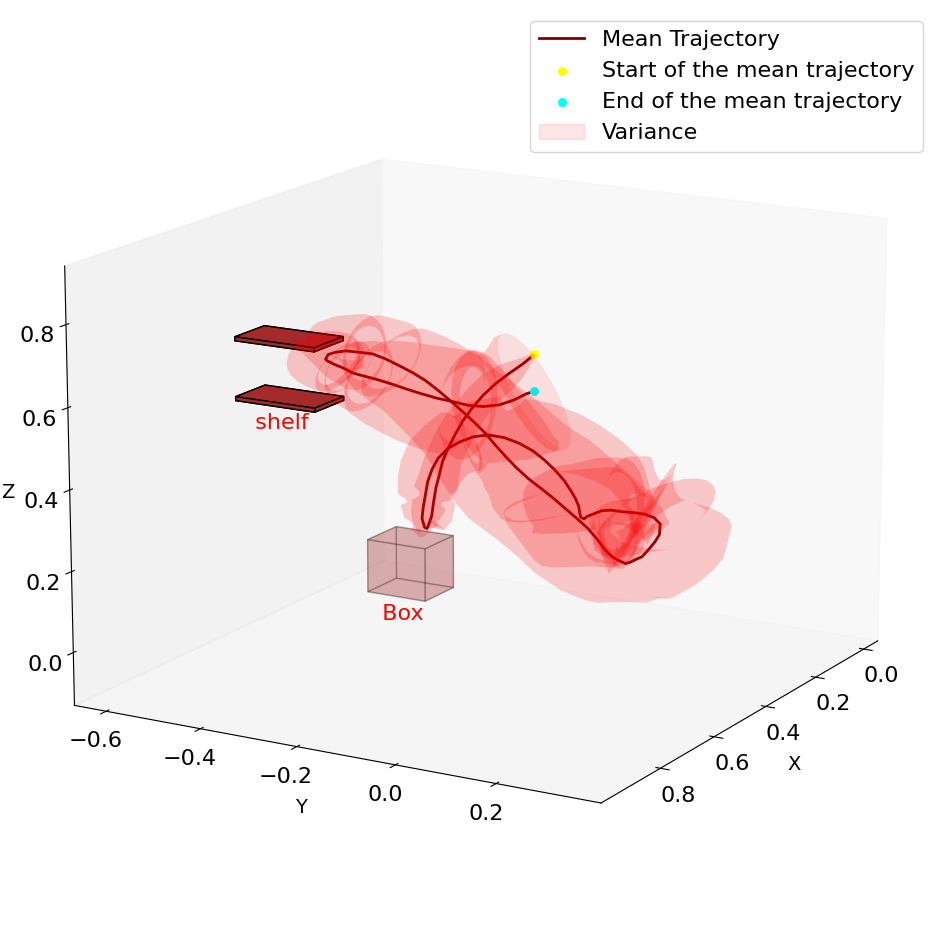}}%
    \\
	\subfloat{\includegraphics[width=0.4\textwidth]{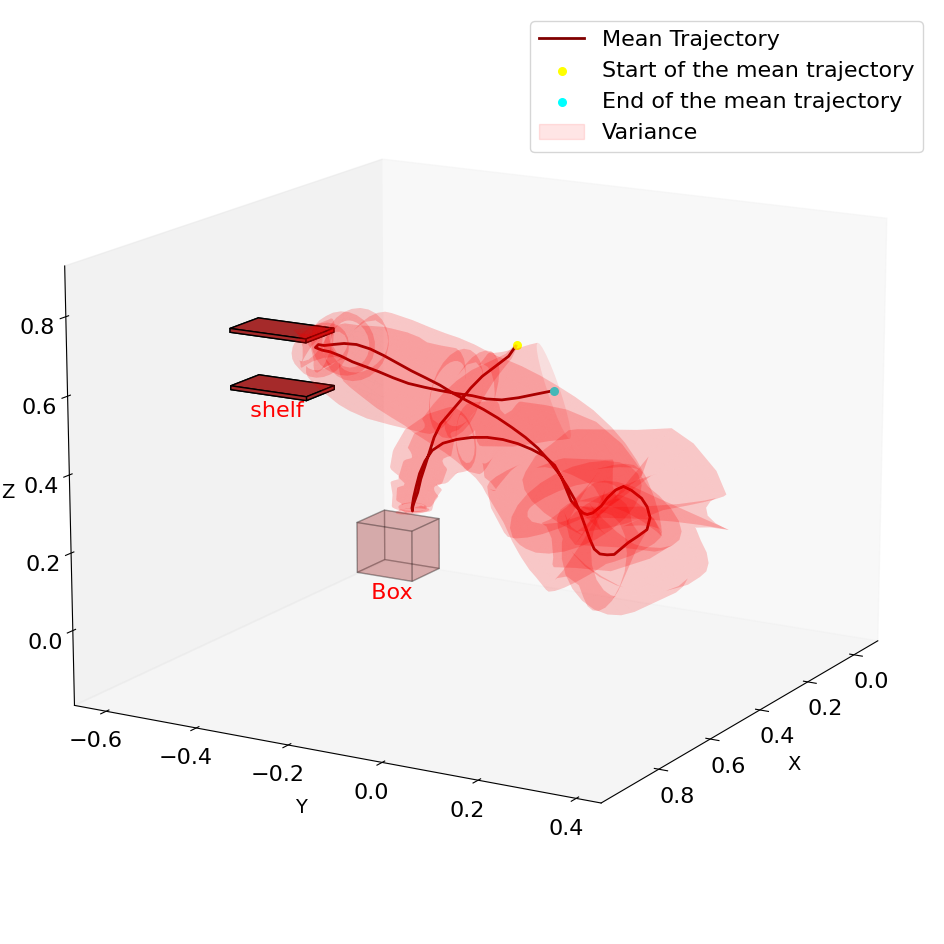}}\\    
	\caption{
		Comparison of task learning methods:
		(top) Mean trajectory and variability across users for Monolithic Policy.
		(bottom) Mean trajectory and variability across demonstrations for \seg Policy.
    }
    \label{fig: learned behaviour}%
\end{figure}

To illustrate the characteristics of the movement models learned using the monolithic and sequential policy teaching approaches, Figure \ref{fig: learned behaviour} (top and bottom) shows the average trajectory and variance of the policies learned using the monolithic and \seg frameworks, providing a visualization of the trajectories' best estimate and their variability across different users.
It can be observed that the learned behaviors from both teaching methods exhibit similar overall patterns, showing that the experiment design successfully achieved a task that provides participants with a \glspl{dof} (segment choices) while limiting the variability between both frameworks, hence yielding a valid direct comparison.

\subsection{Comparison Against Automatic Segmentation}\label{subsec: automatic segmentation}

Automatic segmentation methods rely on users providing a complete task demonstration before it can be segmented, usually using user-provided segmentation labels as a reference for evaluation.
Note that in our setup, users create the segments on the fly (online), and the meaning of each segment is specific to each user, making a direct comparison against an automatically segmented trajectory invalid.

\begin{figure}
	\centering
	\includegraphics[width=0.8\linewidth]{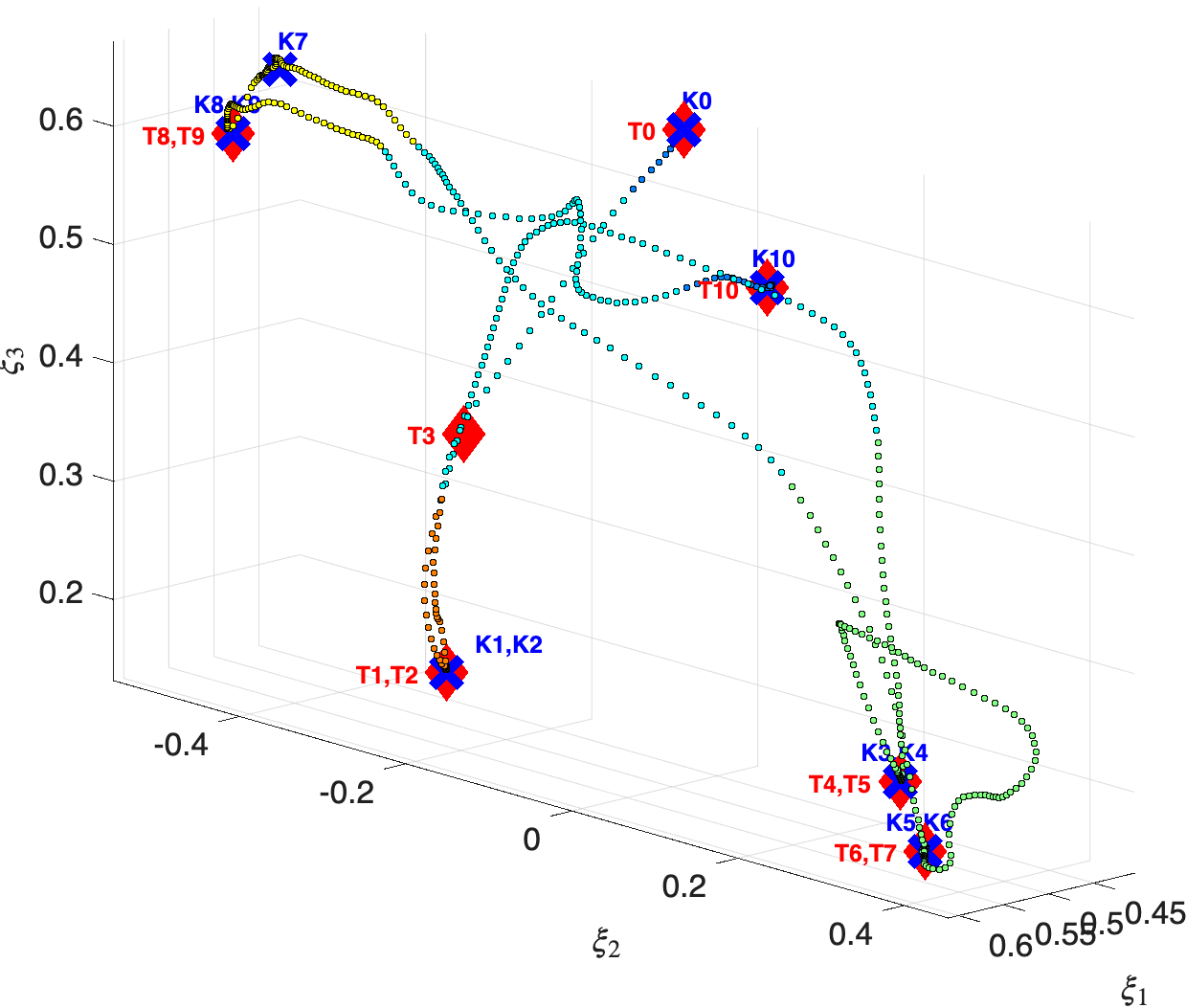}
	\caption{
    Automatic segmentation of user's $\#16$ demonstration using \cite{figueroa2018physically}. Segments (indicated by color) are defined by proximity.
    The user-defined key-frames (blue crosses annotated with $K$) often match the sub-tasks identified in Table \ref{table:tasks} (indicated by diamonds annotated with $\smdpSubtask$).
    }
	\label{fig: seg comparison}
\end{figure}

Furthermore, segmentation methods usually cluster similar points in the trajectory, without a guaranteed physical meaning to the human teacher.
To exemplify this problem, Figure \ref{fig: seg comparison} presents an example trajectory segmented using \cite{figueroa2018physically}, and the respective key-points created by the user (marked with a blue cross ``x''), and the ending key-points of the sub-tasks identified in Table \ref{table:tasks} by us (marked with a red diamond).
Focusing on the automatically generated light-blue segment, one can observe that it contains points from when the robot was moving to fetch the milk carton and when it was placing it on the shelf, creating ambiguities for a policy trained on that segment.
On the other hand, the user's segmentation often matches the task breakdown, highlighting how the user naturally segments the task with intrinsic semantic meaning and associated usefulness in the context of our teaching framework.

\subsection{\gls{fmea} Errors}\label{subsec: fmea errors}

Finally, Table \ref{tab: fmea counts} shows the frequencies of the possible errors highlighted in the \gls{fmea} analysis (Table \ref{table:fmea}) that occurred during the experiments.
The overall percentage of trials having errors is relatively large ($76 \%$ for monolithic and $54.54 \%$ for \seg), highlighting the complexity of the presented long-horizon task and its appropriateness for the study.
The rate for errors related to multiple objects or sub-tasks is higher when using the monolithic method, indicating how focusing on a specific segment improves the teaching signal.
Curiously, using \seg makes users reach the joint limits more often, which we speculate to be caused by disregarding the path taken by the robot to achieve the current pose between segments.

\begin{table}[h]
    \centering
    \caption{Frequency of the errors shown in Table \ref{table:fmea} measured as the rate between the number of occurrences and the total number of trials (\%). Omitted errors did not occur.}
    \label{tab: fmea counts}
    \begin{tabulary}{\linewidth}{c|CCCCCC}
        \toprule
        \textbf{Error Code} & ET01 & ET02 & ET03 & ET04 & ET05 & ET06\\\hline
        \textbf{Monolithic} & $16.27$ & $0$ &  $13.95$&  $18.60$& $6.97$& $4.65$ \\
        \textbf{\seg} & $12.12$ & $0$ &  $3.03$ & $9.09$ & $0$ &  $0$\\\midrule
        
        \textbf{Error Code} & ET07 & ET08 & ET09  & EP01 & EP02 & EG00\\\hline
        \textbf{Monolithic} & $2.32$ & $6.97$ & $0$ & $4.65$ & $2.32$ & $37.2$ \\
        \textbf{\seg} & $6.06$ & $0$ & $6.06$ & $0$ & $0$ &  $12.12$\\
        \bottomrule
    \end{tabulary}
\end{table}

\section{Conclusion}\label{sec: conclusion}

This work presents a structured user study to evaluate human teachers' preferences regarding teaching long-horizon tasks as a monolithic policy or dividing the task into segments.
To evaluate the former, we introduce a novel method called \seg, in which teachers define segmentation points on-the-fly, allowing for progressive task learning, paired with a task designed for testing the methods while yielding consistent policies due to its constrained solution.
Our evaluation indicates a reduction in temporal demand using the proposed method, since it allows users to focus only on the problematic parts of the policy.
From the teaching strategies and preferences, the resulting insights show that some users prefer the proposed method because it provides better control over local correction, despite its increased temporal demand, while others lean towards the monolithic approach due to the simplicity of its single movement demonstration.

Several lessons emerged from our study.
Notably, we observed a wide variability in user behavior and preferences, highlighting the importance of large and diverse participant groups to ensure statistically significant conclusions.
Interestingly, user preference for a teaching method did not align with task success, similar for both methods, suggesting that perceived usability and actual performance may diverge.

\begin{scontents}[store-env=task time]
    Nevertheless, the policies created with \seg are shown to be smoother than the traditional monolithic approach, indicating that users provide better demonstrations and corrections over the segmented trajectories, without affecting the task execution time.
    User-level analysis revealed that the sequential approach improved performance in most cases compared with the monolithic approach.
\end{scontents}

Looking forward, several avenues for future work remain.
First, we plan to explore alternative segmentation interfaces that offer users greater flexibility and control over how sub-tasks are defined.
Second, we aim to extend our evaluation to a broader set of tasks with varying lengths and complexities to assess the generality of our approach.
Finally, we are interested in studying long-term user behavior and co-learning.
Specifically, how teaching efficiency and strategy evolve as users gain more experience with the system and teach multiple, diverse tasks.

\section*{Acknowledgements}
The authors would like to thank Giovanni Franzese and Marta Ferraz for their valuable discussions and assistance at various stages of this work.
LSR was sponsored within the SERICS - SEcurity and RIghts in the CyberSpace and received funding from the European Union Next-Generation EU (Piano Nazionale di Ripresa e Resilienza (PNRR) – Missione 4 Componente 2, Investimento 1.3 – D.D. 1555 11/10/2022, PE00000013, and PE7 - CUP J33C22002810001, D.D. 341 15/03/2022, PE00000014).
RP is sponsored by ARTPARK, IISc Bangalore.
This project is made possible by a contribution from the National Growth Fund program NXTGEN Hightech.
This manuscript reflects only the authors’ views and opinions; neither the European Union nor the European Commission can be considered responsible for them.
This work involved human subjects in its research.
Approval of all ethical and experimental procedures and protocols was granted by TU Delft.

\bibliography{biblio/biblio}  

@article{franzese2023interactive,
	author={Franzese, Giovanni and de Souza Rosa, Leandro and Verburg, Tim and Peternel, Luka and Kober, Jens},
	journal={IEEE/ASME Transactions on Mechatronics}, 
	title={Interactive Imitation Learning of Bimanual Movement Primitives}, 
	year={2024},
	volume={29},
	number={5},
	pages={4006-4018},
	doi={10.1109/TMECH.2023.3295249}
}

@InProceedings{figueroa2018physically,
	title = 	 {A Physically-Consistent Bayesian Non-Parametric Mixture Model for Dynamical System Learning},
	author =       {Figueroa, Nadia and Billard, Aude},
	booktitle = 	 {2nd Conference on Robot Learning},
	pages = 	 {927--946},
	year = 	 {2018},
}

@book{fmea,
	title={The basics of FMEA},
	author={Mikulak, Raymond J and McDermott, Robin and Beauregard, Michael},
	year={2017},
	publisher={CRC press}
}

@article{celemin2022interactive,
  title={Interactive Imitation Learning in Robotics: A Survey},
  author={Celemin, Carlos and P{\'e}rez-Dattari, Rodrigo and Chisari, Eugenio and Franzese, Giovanni and de Souza Rosa, Leandro and Prakash, Ravi and Ajanovi{\'c}, Zlatan and Ferraz, Marta and Valada, Abhinav and Kober, Jens},
  journal={Foundations and Trends{\textregistered} in Robotics},
  volume={10},
  number={1-2},
  pages={1--197},
  year={2022},
  publisher={Now Publishers, Inc.}
}

@book{williams2006gaussian,
	title={Gaussian Processes for Machine Learning},
	author={Williams, Christopher KI and Rasmussen, Carl Edward},
	volume={2},
	number={3},
	year={2006},
	publisher={MIT press Cambridge, MA}
}

@inproceedings{franzese2021ilosa,
	title        = {{ILoSA}: Interactive Learning of Stiffness and Attractors},
	author       = {Franzese, Giovanni and M\'{e}sz\'{a}ros, Anna and Peternel, Luka and Kober, Jens},
	year         = {2021},
	booktitle    = {IEEE/RSJ International Conference on Intelligent Robots and Systems (IROS)},
	pages        = {7778--7785},
	doi          = {10.1109/IROS51168.2021.9636710}
}

@INPROCEEDINGS{mandlekar2020learning, 
	AUTHOR    = {Ajay Mandlekar AND Danfei Xu AND Roberto Martín-Martín AND Silvio Savarese AND Li Fei-Fei}, 
	TITLE     = {{GTI}: Learning to Generalize across Long-Horizon Tasks from Human Demonstrations}, 
	BOOKTITLE = {Robotics: Science and Systems}, 
	YEAR      = {2020}, 
}

@article{tausch2022best,
	title        = {{The best task allocation process is to decide on one’s own: effects of the allocation agent in human–robot interaction on perceived work characteristics and satisfaction}},
	author       = {Tausch, Alina and Kluge, Annette},
	year         = {2022},
	journal      = {{Cogn. Technol. Work}},
	publisher    = {Springer-Verlag},
	address      = {Berlin, Heidelberg},
	volume       = {24},
	number       = {1},
	pages        = {39–55},
	doi          = {10.1007/s10111-020-00656-7},
	issn         = {1435-5558},
	issue_date   = {Feb 2022},
	numpages     = {17},
}

@inproceedings{pantano2023effects,
	title        = {Effects of Robotic Expertise and Task Knowledge on Physical Ergonomics and Joint Efficiency in a Human-Robot Collaboration Task},
	author       = {Pantano, Matteo and Curioni, Arianna and Regulin, Daniel and Kamps, Tobias and Lee, Dongheui},
	year         = {2023},
	booktitle    = {{IEEE-RAS 22nd International Conference on Humanoid Robots (Humanoids)}},
	volume       = {},
	number       = {},
	pages        = {1--8},
	doi          = {10.1109/Humanoids57100.2023.10375163}
}

@inproceedings{welschehold2019combined,
	title        = {Combined Task and Action Learning from Human Demonstrations for Mobile Manipulation Applications},
	author       = {Welschehold, Tim and Abdo, Nichola and Dornhege, Christian and Burgard, Wolfram},
	year         = {2019},
	booktitle    = {IEEE/RSJ International Conference on Intelligent Robots and Systems (IROS)},
	volume       = {},
	number       = {},
	pages        = {4317--4324},
	doi          = {10.1109/IROS40897.2019.8968091},
}

@inproceedings{li_hierarchical_2021,
	title        = {Hierarchical Learning from Demonstrations for Long-Horizon Tasks},
	author       = {Li, Boyao and Li, Jiayi and Lu, Tao and Cai, Yinghao and Wang, Shuo},
	year         = {2021},
	booktitle    = {{IEEE International Conference on Robotics and Automation (ICRA)}},
	pages        = {4545--4551},
	doi          = {10.1109/ICRA48506.2021.9561408}
}

@article{gutzeit2022unsupervised,
	title        = {Unsupervised Segmentation of Human Manipulation Movements Into Building Blocks},
	author       = {Gutzeit, Lisa and Kirchner, Frank},
	year         = {2022},
	journal      = {{IEEE Access}},
	volume       = {10},
	number       = {},
	pages        = {125723--125734},
	doi          = {10.1109/ACCESS.2022.3225914}
}

@phdthesis{fox_bayesian_2009,
	title        = {{Bayesian nonparametric learning of complex dynamical phenomena}},
	author       = {Fox, Emily Beth},
	year         = {2009},
	urldate      = {2023-01-25},
	note         = {Accepted: 2010-05-25T20:43:39Z},
	school  = {Massachusetts Institute of Technology},
	type         = {Thesis}
}

@article{patton2024programming,
	title        = {Programming-by-Demonstration for Long-Horizon Robot Tasks},
	author       = {Patton, Noah and Rahmani, Kia and Missula, Meghana and Biswas, Joydeep and Dillig, I\c{s}\i{}l},
	year         = {2024},
	journal      = {Proc. ACM Program. Lang.},
	volume       = {8},
	number       = {POPL},
}

@inproceedings{mandlekar2023human,
	title        = {Human-in-the-Loop Task and Motion Planning for Imitation Learning},
	author       = {Mandlekar, Ajay and Garrett, Caelan Reed and Xu, Danfei and Fox, Dieter},
	year         = {2023},
	booktitle    = {7th Conference on Robot Learning},
	pages        = {3030--3060},
}

@inproceedings{jiang2024comprehensive,
	author={Jiang, Xinkai and Mattes, Paul and Jia, Xiaogang and Schreiber, Nicolas and Neumann, Gerhard and Lioutikov, Rudolf},
    booktitle={19th ACM/IEEE International Conference on Human-Robot Interaction (HRI)}, 
    title={A Comprehensive User Study on Augmented Reality-Based Data Collection Interfaces for Robot Learning}, 
    year={2024},
    pages={333-342},
}

@INPROCEEDINGS{perez2017c,
	author={Pérez-D'Arpino, Claudia and Shah, Julie A.},
	booktitle={IEEE International Conference on Robotics and Automation (ICRA)}, 
	title={C-LEARN: Learning geometric constraints from demonstrations for multi-step manipulation in shared autonomy}, 
	year={2017},
	volume={},
	number={},
	pages={4058-4065},
	doi={10.1109/ICRA.2017.7989466}}

@inproceedings{mandlekar2021matters,
	title        = {What Matters in Learning from Offline Human Demonstrations for Robot Manipulation},
	author       = {Mandlekar, Ajay and Xu, Danfei and Wong, Josiah and Nasiriany, Soroush and Wang, Chen and Kulkarni, Rohun and Fei-Fei, Li and Savarese, Silvio and Zhu, Yuke and Mart\'in-Mart\'in, Roberto},
	year         = {2022},
	pages        = {1678--1690},
	booktitle    = {5th Conference on Robot Learning},
}

@article{sutton1999mdps,
  title = {Between {MDPs} and Semi-{MDPs}: A Framework for Temporal Abstraction in Reinforcement Learning},
  shorttitle = {Between {MDPs} and Semi-{MDPs}},
  author = {Sutton, Richard S. and Precup, Doina and Singh, Satinder},
  year = {1999},
  journal = {Artificial Intelligence},
  shortjournal = {Artificial Intelligence},
  volume = {112},
  number = {1},
  pages = {181--211},
  issn = {0004-3702},
  doi = {10.1016/S0004-3702(99)00052-1}
}

@article{franzese2025generalizable,
  title={Generalizable Motion Policies Through Keypoint Parameterization and Transportation Maps},
  author={Franzese, Giovanni and Prakash, Ravi and Della Santina, Cosimo and Kober, Jens},
  journal={IEEE Transactions on Robotics},
  year    = {2025},
  volume  = {41},
  number  = {},
  pages   ={4557--4573},
  doi     = {10.1109/TRO.2025.3582821},
}
\bibliographystyle{IEEEtran}

\end{document}